\documentclass[10pt,twocolumn,letterpaper]{article}

\usepackage{cvpr}
\usepackage{times}
\usepackage{epsfig}
\usepackage{graphicx}

\usepackage{amsmath,amssymb,amsfonts,dsfont,bm}
\usepackage{caption}
\usepackage{xcolor}
\usepackage{booktabs, multirow, adjustbox}
\usepackage[pagebackref=true,breaklinks=true,letterpaper=true,colorlinks,bookmarks=false]{hyperref}

\newcommand{\x}{{\rm\bf x}}      
\newcommand{\f}{{\rm\bf f}}      
\newcommand{\h}{{\rm\bf h}}      
\newcommand{\z}{{\rm\bf z}}      
\newcommand{\loss}{{\mathcal L}} 
\newcommand{\E}{{\mathbb E}}     

\newcommand{\wo} {\emph{w/o }}

\cvprfinalcopy 

\def\cvprPaperID{2365} 

\begin{document}

\title{FaceFeat-GAN: a Two-Stage Approach for Identity-Preserving Face Synthesis}
\author{
Yujun Shen\textsuperscript{1}, Bolei Zhou\textsuperscript{1}, Ping Luo\textsuperscript{1,2}, Xiaoou Tang\textsuperscript{1}\\
\textsuperscript{1}CUHK - SenseTime Joint Lab, The Chinese University of Hong Kong\\
\textsuperscript{2}Shenzhen Institutes of Advanced Technology, Chinese Academy of Sciences\\
{\tt\small \{sy116, bzhou, pluo, xtang\}@ie.cuhk.edu.hk}
}

\maketitle

\begin{abstract}
The advance of Generative Adversarial Networks (GANs) enables realistic face image synthesis.
However, synthesizing face images that preserve facial identity as well as have high diversity within each identity remains challenging.
To address this problem, we present FaceFeat-GAN, a novel generative model that improves both image quality and diversity by using two stages.
Unlike existing single-stage models that map random noise to image directly, our two-stage synthesis includes the first stage of diverse feature generation and the second stage of feature-to-image rendering. 
The competitions between generators and discriminators are carefully designed in both stages with different objective functions.
Specially, in the first stage, they compete in the feature domain to \emph{synthesize various facial features} rather than images.
In the second stage, they compete in the image domain to render photo-realistic images that contain high diversity but preserve identity.
Extensive experiments show that FaceFeat-GAN generates images that not only retain identity information but also have high diversity and quality, significantly outperforming previous methods.
\end{abstract}

\section{Introduction}\label{sec:introduction}
Generative Adversarial Networks (GANs) make a significant progress to face synthesis, leading to a great number of applications such as face editing \cite{fadergan}, face recognition \cite{dagan}, and face detection \cite{find_tiny_faces_gan}.
An image synthesis model is commonly evaluated by two criteria.
The first one is \emph{image quality}, which measures how realistic the generated images are compared to the real one.
The second one is \emph{image diversity}, which measures the variations of the synthesized contents.
A key challenge is to balance these two criteria and produce images that are both photo-realistic and of large variety.
Although the advance of GANs has led to significant breakthroughs in unconstrained face image synthesis \cite{wgan,began,progressivegan}, this challenge remains unsolved in the case of generating identity-preserving faces.

\begin{figure}[t]
  \centering
  \includegraphics[width=0.9\linewidth]{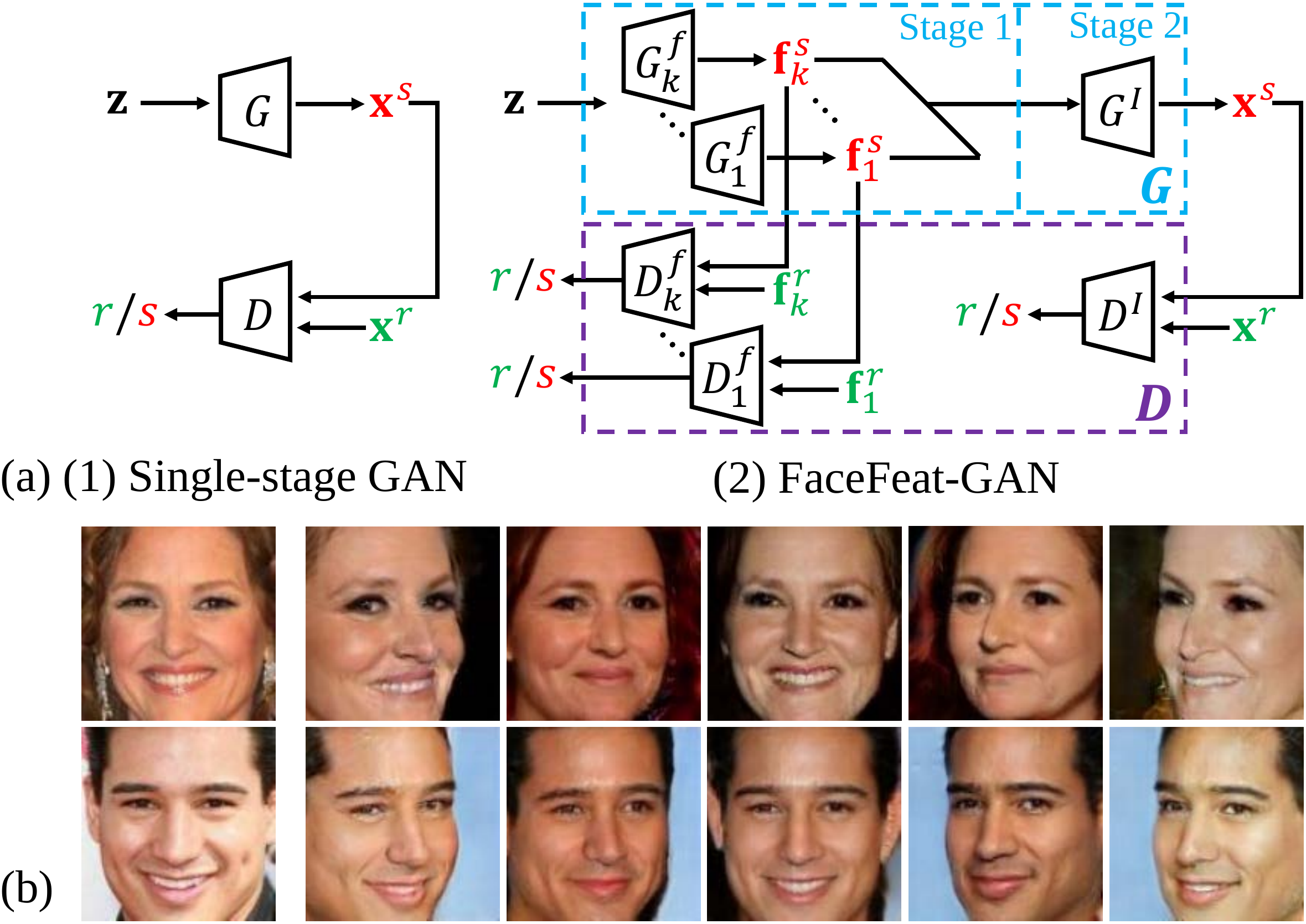}
  \caption{\small{
    Compared to the conventional single-stage GAN in (a.1), we propose FaceFeat-GAN in (a.2) with a two-stage generator $G$.
    First stage generates a collection of facial features with respect to various attributes, such as poses and expressions, while the second stage takes these features as input and then renders photo-realistic face images.
    Correspondingly, the discriminator $D$ also has a two-level competition with $G$ in both feature domain and image domain.
    (b) visualizes some samples generated by FaceFeat-GAN, which are of high diversity as well as preserve the person identity. The first column is the reference image, and the other columns are results synthesized by FaceFeat-GAN.
  }}
  \label{fig:attraction}
\end{figure}

As shown in Fig.\ref{fig:attraction}(a.1), conventional single-stage GAN model is formulated as a two-player game between a discriminator $D$ and a generator $G$.
By competing with $D$, $G$ is eventually able to synthesize images $\x^s$ that are as realistic as real ones $\x^r$.
However, the situation becomes more complex when a constraint is imposed to the above generation process, such as preserving the identity of a face image.
There are two main difficulties in this conditional generative problem.
One is \emph{how to extract and convert identity information to high-quality face images (\eg sharpness, identity)}, while the other is \emph{how to increase the image diversity (\eg viewpoint variations) of the synthesized faces of the same identity}.
Solving them simultaneously needs a trade-off.

One intuitive solution is to feed the identity label to the generator $G$ to guide the synthesis process \cite{infogan}.
However facial identity is so complex that only using a label as supervision is not enough for $G$ to learn the identity information and achieve high-quality synthesis.
Accordingly, several work \cite{ffgan,tpgan} handled the sparse supervision by introducing pixel-wise supervision with paired training data.
Each pair contains two images of the same identity, one as input and the other as the target output.
For example, an image with canonical viewpoint is treated as supervision of $G$ to alleviate the training difficulty of face frontalization task.
However, this kind of per-pixel supervision severely limits the image diversity, because $G$ would only generate one desired output for each input in order to minimize the pixel-wise loss.
In this regard, these models were usually designed for the tasks with single-mode output, such as style transfer \cite{face_style_transfer}.
%

In this work we propose a novel two-stage generative model, termed FaceFeat-GAN, to tackle the trade-off of the aforementioned two criteria in a unified framework.
We divide face synthesis into two stages, where the first stage accounts for synthesis diversity by producing various facial features, while the second stage further renders high-quality identity-preserving face image with the above generated features.

As shown in Fig.\ref{fig:attraction}(a.2), in the first stage, we employ a series of feature generators, $\{G_i^f\}_{i=1}^k$, to generate a set of diverse facial features $\{\f_i^s\}_{i=1}^k$.
Here $k$ is the number of feature generators, each of which produces the feature corresponding to a particular facial attribute, such as pose \cite{drgan}, expression \cite{dyadgan}, age \cite{aginggan}, \etc.
Note that \emph{synthesizing semantic features as the intermediate step} for the later image synthesis is a key contribution distinguishing FaceFeat-GAN from prior work.
In the second stage, an image generator $G^I$ takes all these features as inputs and outputs a photo-realistic face image.
The pixel-wise supervision can be easily applied to this stage without affecting the diversity of the first stage, since $G^I$ only focuses on learning the mapping from feature space to image space regardless of whether the features are real or fake.
In addition, for each generator, both $G_i^f$ and $G^I$, we introduce a discriminator to ensure the realness of synthesized results, forming a two-level competition.
In other words, $G_i^f$ competes with $D_i^f$ in the semantic feature domain to synthesize face features, while $G^I$ competes with $D^I$ in image domain to produce face images.

FaceFeat-GAN has two advantages compared to existing methods.
First, benefiting from the competition in feature space, the facial features synthesized by various feature generators significantly improve image diversity of the same identity (see Fig.\ref{fig:attraction}(b)).
Second, mapping facial features to image space naturally encodes identity information to achieve high-quality identity-preserving face synthesis.

This work has three \textbf{contributions}:
(1) We propose an effective two-stage generation framework.
Instead of training independently, these two stages collaborate with each other through a carefully-designed two-level competition by GANs.
(2) FaceFeat-GAN finds a good way to deal with the trade-off between image quality and diversity in conditional generative problem.
(3) Extensive experiments show that FaceFeat-GAN synthesizes identity-preserving images that are both photo-realistic and highly-diverse, surpassing previous work.


\section{Related Work}\label{sec:realted work}
\noindent\textbf{Face Representation.}
Learning facial features has been extensively applied to face-related tasks, such as face recognition \cite{sphereface}, face alignment \cite{face_alignment}, and 3D face reconstruction \cite{3d_reconstruction}.
Recent work \cite{sdgan,d2ae} has demonstrated the great potential of learning disentangled features from face images, making it possible to encode all information of a face image in a complete feature space and manipulate them independently.

Some previous work employed facial features for face synthesis.
DR-GAN \cite{drgan} used pose code to adjust head pose, FaceID-GAN \cite{faceidgan} used expression features to modify face expression, and face contour was used by \cite{semantic_manipulation_gan} to manipulate facial shape.
However, all features in these works are manually specified, limiting their authenticity and variety.
On the contrary, FaceFeat-GAN employs feature generators to produce features, which are learned from real feature distributions.
In this way, these synthesized features are more realistic and also have higher diversity.

Besides the features mentioned above, some work \cite{dagan, ffgan, 3dpim} introduced 3D information to assist face synthesis.
3D Morphable Model (3DMM) \cite{3dmm} is a commonly used model.
It represents a 3D face with a set of bases and builds a bridge between 3D face and 2D image with a series of transformation parameters, making it suitable to describe the expression and pose of a face image.
In this work, we use 3DMM parameters as the pose and expression feature.

\noindent\textbf{Identity-Preserving Face Synthesis.}
Generative Adversarial Network (GAN) \cite{gan} is one of the most powerful models for face synthesis.
It consists of a generator $G$ and a discriminator $D$ that compete with each other, formulating a two-player game.
When adding the constraint of preserving identity to the original generative problem, it is a common practice to pass the identity information, $\ell_{id}$, to the generator $G$ and also use $\ell_{id}$ as supervision.
Prior work tried various forms of information for identity representation, such as identity label \cite{infogan} and identity feature \cite{drgan}, but all of them suffer from incomplete identity maintenance.
To solve this problem, FaceID-GAN \cite{faceidgan} proposed a three-player competition where the generator $G$ not only competes with the discriminator $D$ from image quality aspect, but also competes with an identity classifier $C$ from identity preservation aspect.
However, the image quality is still not as satisfying as methods which employ a ground truth image to guide the generation process \cite{tpgan,ffgan,pim,3dpim}.
In general, the ground truth image can tell the generator what value should be produced for each pixel, which is a much more accurate supervision.
On the other hand, however, the pixel-wise supervision leads to extremely low diversity, since the target output for each input is fixed.
That is the reason why these models are always designed for many-to-one mapping task, such as face frontalization.

Variational Auto-Encoder (VAE) \cite{vae} is another kind of generative model.
The key idea is to learn a continuous latent feature space with an auto-encoder structure, such that each sample in the latent space can be decoded to a realistic image.
Some work \cite{attribute2image} introduced identity constraint to VAE for identity-preserving face synthesis, but the produced images suffer from blurring as it lacks a discriminator to compete with the image decoder.
CVAE-GAN \cite{cvaegan} attempted to tackle the blurring problem by combining GAN and conditional VAE together.
Based on auto-encoder structure, the above methods included pixel-wise supervision automatically.
Nevertheless, the decoder aims at reconstructing the input image regardless of the input randomness, which may cause some ambiguities and restrict the diversity of the generation results.
Instead, \cite{opensetgan} proposed a feasible solution by using different attribute images as target output images corresponding to different input noises.
However, the attribute images are not always with the same identity as input image.
Using them as supervision will lead to identity information loss.

In contrast, FaceFeat-GAN solves the above problems with two stages.
The first stage produces synthesized features by learning the distribution of real features that are extracted from real images, to enhance diversity.
The second stage learns a mapping from feature space to image space by reconstructing the input image with both per-pixel supervision and adversarial supervision, to improve image quality and preserve identity.
In other words, as long as the features produced by the first stage is real enough, generator of the second stage will be able to decode them to photo-realistic images.
In this way, the two stages focus on different aspects, but collaborate together for better synthesis.

There are also classic methods  \cite{rotating_your_face,hpen,synthesizing_normalized_faces} that achieved identity-preserving face synthesis without using generative models.
We would like to acknowledge their contributions.

\noindent\textbf{Multiple Competitors.} 
In the GAN literature, there are some models with multiple competitors.
For example, multiple generators are used in \cite{multi-generator_gan} to solve mode collapse problem.
Some work \cite{multi-adversarial_gan,dual_discriminator_gan} used two or more discriminators to improve the differential ability so that the generator can produce more realistic images.
Several models \cite{progressivegan,semantic_manipulation_gan} established competition between generator $G$ and discriminator $D$ under different spatial resolutions to improve image quality.
\cite{triplegan,faceidgan} trained $G$ by competing not only with the discriminator $D$ but also with a classifier $C$ to better solve conditional generative problem.
Similarly, the discriminator in \cite{stargan} was treated as a domain classifier to achieve across-domain synthesis.
Different from them, however, FaceFeat-GAN presents a two-level competition from both high-level feature domain and low-level image domain, which is more effective than prior work.
In addition, the competitions in these two domains are not independent from each other, but collaborate to achieve better results.

\begin{figure*}[t]
  \centering
  \includegraphics[width=0.95\linewidth]{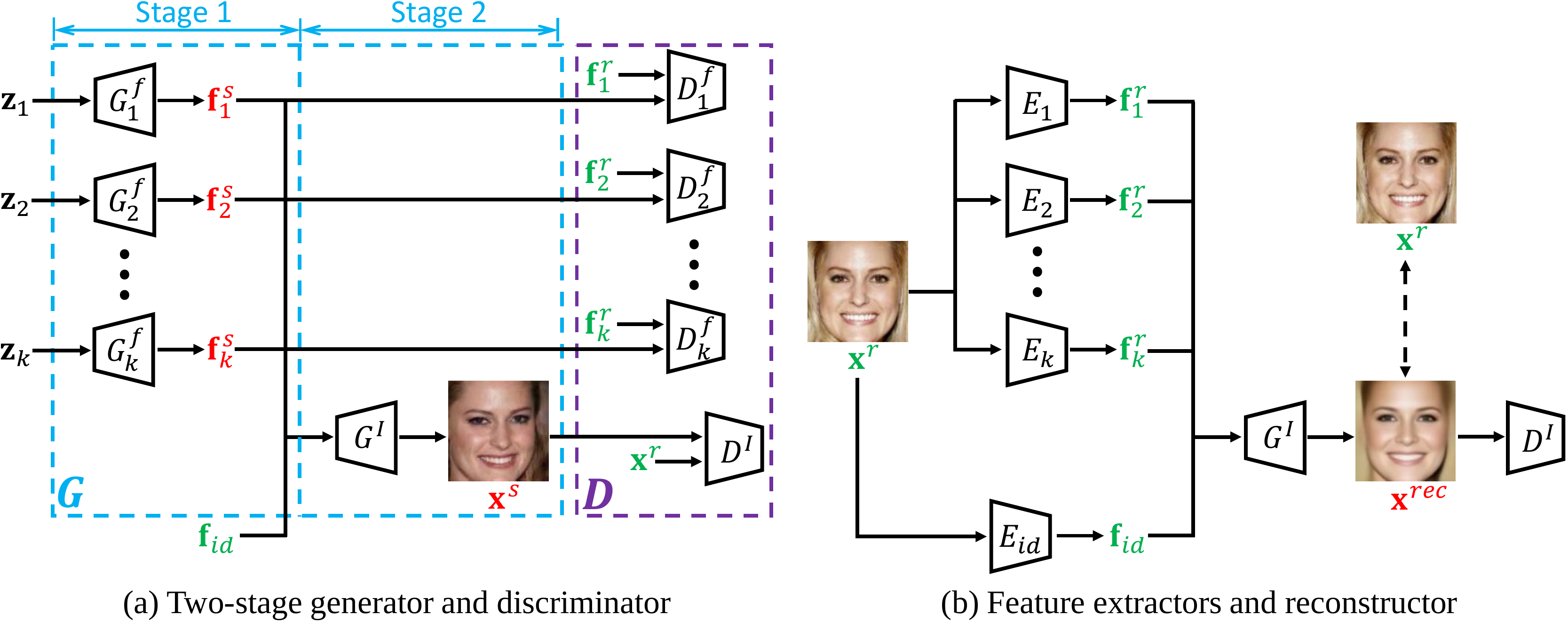}
  \caption{\small{
    (a) illustrates the framework of the FaceFeat-GAN. It consists of a two-stage generator $G$, which generates facial features in the first stage and synthesizes high-quality face images from these features in the second stage, and a discriminator $D$, which competes with $G$ from both high-level feature domain and low-level image domain.
    To preserve identity information, $G^I$ also takes the identity feature $\f_{id}$ as reference, which is extracted by a face recognition module $E_{id}$, as in (b).
    (b) shows that besides synthesizing new images, $G^I$ is also trained to reconstruct the input $\x^r$ with real features extracted from it.
    The dashed two-way arrow indicates the pixel-wise supervision.
    Better viewed in color.
  }}
  \label{fig:framework}
\end{figure*}

\section{FaceFeat-GAN}\label{sec:method}
\noindent\textbf{Overview.}
Fig.\ref{fig:framework} outlines our framework.
Like existing GAN models, FaceFeat-GAN is formulated as a competition between generators and discriminators.
However, we design a more delicate competition strategy by dividing the synthesis process into two stages, as shown in Fig.\ref{fig:framework}(a), including 
(1) producing realistic but diverse facial features $\{\f_i^s\}_{i=1}^k$ from random noises $\{\z_i\}_{i=1}^k$ using a series of feature generators $\{G_i^f\}_{i=1}^k$,
and (2) decoding the above features to a synthesized image $\x^s$ with the image generator $G^I$.
Besides generated features $\{\f_i^s\}_{i=1}^k$, $G^I$ also takes the identity feature $\f_{id}$ to gain identity information.
To guarantee the realness of the synthesized results of each stage, we introduce feature discriminators $\{D_i^f\}_{i=1}^k$ and image discriminators $D^I$ to compete with $\{G_i^f\}_{i=1}^k$ and $G^I$ respectively, forming a two-level competition.

Unlike conventional GAN, FaceFeat-GAN not only generates fake images, but also reconstructs the input image to acquire more accurate pixel-wise supervision, as shown in Fig.\ref{fig:framework}(b).
Specifically, a series of feature extractors $\{E_i\}_{i=1}^k$ are employed to extract real features $\{\f_i^r\}_{i=1}^k$ from input image $\x^r$, and an face recognition module $E_{id}$ is used to extract identity feature $\f_{id}$.
Then the same $G^I$ as above takes these features as inputs and produces $\x^{rec}$ to reconstruct $\x^r$.
Here, $\x^{rec}$ will also be treated as fake image by $D^I$.
Besides $G^I$ can learn a better mapping from feature space to image space under the identity-preserving constraint, another advantage in doing so is that the real features $\f_i^r$ extracted by $E_i$ can be used as a reference for $D_i^f$ to make $G_i^f$ produce more realistic features.

\noindent\textbf{Loss Functions.}
To summarize, the objective functions for $\{G_i^f\}_{i=1}^k$ and $\{D_i^f\}_{i=1}^k$ are as follows
\begin{align}
  & \min_{\Theta_{G_i^f}}\loss_{G_i^f}=\phi_i^f(\f_i^s)+\lambda_i^f\phi^I(\x^s),~i=1\dots k, \label{eq:loss_gf} \\
  & \min_{\Theta_{D_i^f}}\loss_{D_i^f}=\phi_i^f(\f_i^r)-\phi_i^f(\f_i^s),~i=1\dots k,        \label{eq:loss_df}
\end{align}
while $G^I$ and $D^I$ are trained with
\begin{align}
  &\begin{aligned}
    \min_{\Theta_{G^I}}\loss_ {G^I}=&\ \phi^I(\x^{rec})+\phi_{rec}(\x^r, \x^{rec})  \\
                                    &+\lambda_1\phi_{id}(\x^{rec})+\lambda_2\phi_{id}(\x^s),
   \end{aligned}                                                                            \label{eq:loss_gi} \\
  &\min_{\Theta_{D^I}}\loss_{D^I}=\phi^I(\x^r)-\lambda_3\phi^I(\x^s)-\lambda_4\phi^I(\x^{rec}), \label{eq:loss_di}
\end{align}
where $\phi_{rec}(\x^r, \x^{rec})=||\x^r-\x^{rec}||_1$ is the $l_1$ reconstruction loss, and $\phi_{id}(\cdot)$ is the loss function to measure identity-preserving quality.
In addition, $\phi_i^f(\cdot)$ is the energy function to determine whether a facial feature is from real domain or fake domain.
Similarly, $\phi^I(\cdot)$ is the energy function to determine whether an image is real or synthesized.
We have $\lambda_i^f$, $\lambda_1$, $\lambda_2$, $\lambda_3$, and $\lambda_4$ denoting the strengths of different terms.
More details will be discussed in the following sections.

\subsection{Feature Extractors}\label{subsec:feature extraction}
According to Fig.\ref{fig:framework}(b), there are $k$ feature extractors $\{E_i\}_{i=1}^k$ in addition with a face recognition engine $E_{id}$ to extract identity feature.
Among them, each $E_i$ represents for a facial feature corresponding to a particular face attribute.
In the following experiment, we let $k=2$, but the framework is flexible to include more facial features.
More specifically, we use a 3DMM feature $\f_1^r=E_1(\x^r)$ to model pose and expression, and a general feature $\f_2^r=E_2(\x^r)$ to represent other facial variations.

\vspace{5pt}
\noindent\textbf{Identity Feature $E_{id}(\x^r)$.}
To encode identity information, we introduce a face recognition module to extract \emph{identity feature} $\f_{id}$ from input image $\x^r$.
This model is trained as a classification task with cross-entropy loss
\begin{align}
\min_{\Theta_{E_{id}}}\loss_{E_{id}}=\sum_{j=1}^N-\{\ell^r_{id}\}_j\log(\{\sigma({\rm\bf W}_{id}^T\f_{id}+{\rm\bf b}_{id})\}_j), \label{eq:loss_eid}
\end{align}
where ${\rm\bf W}_{id}$ and ${\rm\bf b}_{id}$ are the weight and bias parameters of the fully-connected layer following feature $\f_{id}$ and $\sigma(\cdot)$ indicates the softmax function.
$\ell_{id}^r$ is the ground truth identity label of image $\x^r$ and $N$ is the total number of subjects.

\vspace{5pt}
\noindent\textbf{3DMM Feature $E_1(\x^r)$.}
3D Morphable Model \cite{3dmm} is able to describe a 2D image in 3D space with a set of shape basis ${\rm\bf A}_{id}$ \cite{3dmm_shape} and another set of expression basis ${\rm\bf A}_{exp}$ \cite{3dmm_exp}, making it suitable for \emph{pose and expression} representation.
Usually, 3DMM is formulated as
\begin{equation}\label{eq:3dmm}
  \begin{split}
  &{\rm\bf S}={\rm\bf\overline S}+{\rm\bf A}_{id}{\bm\alpha}_{id}+{\rm\bf A}_{exp}{\bm\alpha}_{exp},\\
  &{\rm\bf s}= f{\rm\bf R}(\alpha,\beta,\gamma){\rm\bf S}+{\rm\bf t},\\
  &\f_{3d}={[{{\bm\alpha}_{id}}^T,{{\bm\alpha}_{exp}}^T,f,\alpha,\beta,\gamma,{\rm\bf t}^T]}^T,\\
  \end{split}
\end{equation}
where ${\rm\bf\overline S}$ is the mean shape.
${\bm\alpha}_{id}$ and ${\bm\alpha}_{exp}$ are the coefficients corresponding to ${\rm\bf A}_{id}$ and ${\rm\bf A}_{exp}$ respectively.
Furthermore, $f$, ${\rm\bf R}(\alpha,\beta,\gamma)$, and ${\rm\bf t}=[t_x,t_y, t_z]^T$ are scaling coefficients, rotation matrix, and translation coefficients, which are used for projecting the face from 3D coordinate system {\rm\bf S} back to image coordinate system {\rm\bf s}, and $\f_{3d}$ is the complete 3DMM parameters.
Following \cite{hpen,3ddfa}, the ground truth parameters $\f_{3d}^{gt}$ can be estimated off-line, and then they are learned by using a Convolutional Neural Network (CNN) with loss function
\begin{align}
  \min_{{\Theta}_{E_1}}\loss_{E_1}={(E_1(\x^r)-\f_{3d}^{gt})}^T{\rm\bf W}_{3d}(E_1(\x^r)-\f_{3d}^{gt}), \label{eq:loss_e3d}
\end{align}
where ${\rm\bf W}_{3d}$ is a diagonal matrix, where each value indicates the importance of a particular element in $\f_{3d}$.

\vspace{5pt}
\noindent\textbf{General Feature $E_2(\x^r)$.}
Only having identity, pose, and expression features is not sufficient to describe a face.
Accordingly, we use an additional encoder to learn a more \emph{general feature} to realize complete representation.
This is achieved by trying to reconstruct the input face with the help of $G_I$.
In this way, $E_2$ is trained with
\begin{align}
  \min_{{\Theta}_{E_2}}\loss_{E_2}=||\x^r-G^I(\f_{id},\f_1^r,E_2(\x^r))||_1. \label{eq:loss_egen}
\end{align}

\subsection{Two-Stage Face Generation}\label{subsec:two-stage generation}
As mentioned before, instead of generating images directly, FaceFeat-GAN presents a two-stage generation, with the first stage to synthesize diverse features, and the second stage to decode the features to high-quality images.

\vspace{5pt}
\noindent\textbf{Stage-1: Feature Generation.}
With lower dimension compared to image, feature is much easier to generate.
We incorporate a GAN model for each facial feature except identity, and attempt to generate features with different semantic meanings independently.
Both the generators $G_i^f$ and discriminators $D_i^f$ employ Multi-Layer Perceptron (MLP) structures.
Each $D_i^f$ tries to distinguish real features $\f_i^r$ from fake features $\f_i^s$ that are synthesized by $G_i^f$.
This is treated as a binary classification problem.
Given a feature $\f_i$, $D_i^f$ will output the probability of it belonging to the real domain and is trained with Eq.\eqref{eq:loss_df}.
We have
\begin{align}
  \phi_i^f(\f_i)=-\E_{\f_i\sim P_{\f_i}}[\log(D_i^f(\f_i))], \label{eq:phi_f}
\end{align}
where $P_{\f_i}$ is the distribution to which $\f_i$ is subjected.
Meanwhile, $G_i^f$ tries to fool $D_i^f$ with the opposite objective functions, as shown in the first term of Eq.\eqref{eq:loss_gf} and the second term in Eq.\eqref{eq:loss_df}.

\vspace{5pt}
\noindent\textbf{Stage-2: Image Generation.}
To synthesize identity-preserving faces,
we introduce an image generator $G^I$ to map features to image space after all features are prepared in stage-1.
Similarly, we use an image discriminator $D^I$, which is a CNN, to determine whether an image $\x$ is real or synthesized using the following energy function
\begin{align}
  \phi^I(\x)=-\E_{\x\sim P_{\x}}[\log(D^I(\x))], \label{eq:phi_i}
\end{align}
where $P_{\x}$ is the distribution of image space with respect to $\x$.

As shown in Fig.\ref{fig:framework}, $G^I$ not only synthesizes a new image $\x^s=G^I(\f_{id},\f_1^s,\f_2^s)$ with features generated by $G_1^f$ and $G_2^f$, but also reconstructs $\x^r$ by producing $\x^{rec}=G^I(\f_{id},\f_1^r,\f_2^r)$ with real features extracted by $E_1$ and $E_2$.

With the reconstruction loss, as the second term in Eq.\eqref{eq:loss_gi}, $G^I$ is able to learn a better mapping from feature space to image space.
Besides using the pixel-wise supervision, we also have $D^I$ with the purpose to determine $\x^{rec}$ as fake, shown as the third term in Eq.\eqref{eq:loss_di}, forcing $G^I$ to improve the decoding ability.
Moreover, to maintain identity, we desire the identity features of $\x^{rec}$ and $\x^{s}$ to be as close to $\f_{id}$ as possible.
Therefore, $G^I$ is also trained with the last two terms in Eq.\eqref{eq:loss_gi}, and we have
\begin{align}
  \phi_{id}(\x)=||\f_{id}-E_{id}(\x)||_2^2. \label{eq:phi_id}
\end{align}

\vspace{5pt}
\noindent\textbf{Two-level Competition.}
The above two stages work collaboratively.
From Eq.\eqref{eq:loss_gi} we see that $G^I$ competes with $D^I$ by minimizing $\phi^I(\x^{rec})$ (the third term Eq.\eqref{eq:loss_di}), but not $\phi^I(\x^{s})$ (the second term in Eq.\eqref{eq:loss_di}).
This is because the latter competition is taken over by the feature generators $\{G_i^f\}_{i=1}^k$, which can be seen in the second term in Eq.\eqref{eq:loss_gf}.

There are two advantages in doing so.
On one hand, $G^I$ can focus on learning the feature-to-image mapping.
It may cause some ambiguity to $G^I$ if it is also required to improve synthesis process with features $\{\f_i^s\}_{i=1}^k$, which are produced by some isolated networks $\{G_i^f\}_{i=1}^k$.
On the other hand, $\{G_i^f\}_{i=1}^k$ are able to learn more realistic features with the competitions from not only feature domain but also image domain.
In this way, both $\{G_i^f\}_{i=1}^k$ and $G^I$ are able to do their best with respective purposes in this two-stage generation.

\subsection{Attribute Interpolation and Manipulation}\label{subsec:attribute control}
Besides generating highly-diverse identity-preserving faces, FaceFeat-GAN is also able to manipulate the attributes of the synthesized image independently.
This benefits from the $k$ irrelevant feature generators $\{G\}_{i=1}^k$.

More specifically, after the entire model converges, we generate a bunch of images by randomly sampling from each noise space.
Then, we explore the relationship between the input random noise and the corresponding attribute by annotating the output faces.
With this information, we can manipulate the generation process by feeding the network with specific noise.
For example, suppose we have noise $\z_1^1$ representing for left viewpoint and $\z_1^2$ for right viewpoint, then using interpolations between $\z_1^1$ and $\z_1^2$ as inputs will be able to produce face with arbitrary viewpoint.
Furthermore, other attributes will remain unaffected as long as other noises are kept the same.

\section{Experiments}\label{sec:experiments}
FaceFeat-GAN aims at synthesizing identity-preserving face images with both high quality and high diversity.
We design various experiments from these three aspects, including identity-preserving capacity, image quality, and image diversity, to evaluate its performance and compare it with existing methods.

\noindent\textbf{Datasets.}
We briefly introduce the datasets used in this work.
\textbf{CASIA-WebFace}, consisting of 494,414 images of 10,575 subjects \cite{casia}, is one of the most widely used datasets for face recognition.
This work treats it as the training set.
\textbf{LFW}, which contains 13,233 images of 5,749 subjects collected in the wild  \cite{lfw}, is a popular benchmark for face recognition.
We use it as a validtion set to evaluate the identity-preserving property of FaceFeat-GAN, similar as existing work \cite{ffgan,tpgan,faceidgan}.
\textbf{IJB-A} constains 25,808 images of 500 subjects \cite{ijba}.
We remove the 26 overlapping subjects between CASIA-WebFace and IJB-A at the training stage, and use it to further evaluate the identity-preserving property.
\textbf{CelebA} is a large-scale dataset that contains 202,599 images of 10,177 subjects \cite{celeba}.
We also treat is as the test set to compare with other start-of-the-art methods from both image quality and image diversity. In addition, following \cite{faceidgan}, we train a deep face recognition model on the \textbf{MS-Celeb-1M} dataset \cite{ms1m}.
This model is used for computing the identity similarity between two images and is independent from this work.
%

\noindent\textbf{Implementation details.}
In this work, both input and output images are of size $128\times128$, and all input faces are aligned by using \cite{face_alignment}.
$E_{id}$ employs ResNet-50 structure \cite{resnet} to extract identity features $\f_{id}$, $E_1$ employs ResNet-18 structure to extract 3DMM features $\f_1^r$ from real input images, and $E_2$ employs the encoder structure in BEGAN \cite{began} to extract general features $\f_2^r$.
Among them, $\f_{id}$ is a 256d vector, while $\f_1^r$ and $\f_2^r$ are 30d and 256d vectors respectively.
Here, only ${\bm\alpha}_{exp}$ and $\beta$ in Eq.\eqref{eq:3dmm} are used as $\f_1$.
We also fix the $l_2$-norm of both $\f_{id}$ and $\f_2$ to be 64 when training $E_{id}$, and then re-normalize it to 16 before feeding them into $G^I$.
Each feature generator, \ie $G_1^f$ and $G_2^f$, takes a 64d vector subject to uniform distribution on $[-1,1]$ as input, and employs a four-layer MLP structure with numbers of hidden neurons to be $[128, 256, 256]$.
$D_1^f$ and $D_2^f$ also use four-layer MLP structures with $[256, 256, 128]$ hidden neurons.
As for the image generator-discriminator pair, $G^I$ and $D^I$ apply the structures described in BEGAN.

The loss weights $\lambda_i^f$, $\lambda_1$, $\lambda_2$, $\lambda_3$, and $\lambda_4$ are set to make the corresponding terms numerically comparable such that no loss function will dominate the training process.
Before training, $E_{id}$ and $E_1$ are pre-trained with identity label and 3DMM ground truth $\f_{ed}^{gt}$ respectively, to alleviate the training difficulty of other components.
During the training process, no additional annotations are required.
All parts of FaceFeat-GAN apply Adam optimizer \cite{adam} with an initial learning rate $8e^{-5}$, and the learning rate decays to $5e^{-5}$ at the $100k$-th step.
The whole network is updated with $200k$ steps with batch size 96.

\subsection{Identity-Preserving Capacity}\label{subsec:identity-preserving property}
In this part, we validate the identity-preserving capacity of FaceFeat-GAN.
To measure the identity similarity between the real images and the generated ones, we train a face recognition model on MS-Celeb-1M dataset to extract identity features from faces, where the training data are totally independent from FaceFeat-GAN.
Then, a similarity score is computed by using cosine distance as the metric between two extracted features.
This model achieves $(93.4\pm0.5)\%$ face verification accuracy at FAR 0.001 on IJB-A benchmark, making the scores convincing.

We evaluate FaceFeat-GAN on two most frequently used face verification benchmarks, \ie LFW and IJB-A, without fine-tuning the model on these datasets.
First, we generate one image for each image in LFW by using FaceFeat-GAN.
We then test the face verification accuracy on the generated images following previous work \cite{ffgan,faceidgan}.
According to the results shown in Tab.\ref{tab:lfw}, we see that our work surpasses the state-of-the-art methods, indicating that FaceFeat-GAN better preserves identity information.

Second, we design a particular experiment on IJB-A dataset to test whether FaceFeat-GAN can generate diverse faces while retaining the identity simultaneously.
Unlike the other benchmarks, IJB-A defines template matching for face verification, where each facial template contains various amount of images with the same identity.
Inspired by this process, we establish each template by using both real and synthesized data, and evaluate the verification and identification performance by gradually adjusting the ratio of synthesized data from 0\% to 100\%.

As shown in Tab.\ref{tab:ijba}, results by only using synthesized data (last row) is almost as good as those by only using real data (first row), implying that all generated faces from one identity are still of the same identity.
The results with 50\% synthesized data (third row) is also very impressive, indicating that the distributions of real images and synthesized images are close to each other with respect to face identity.
This benefits from the two-stage generation that is able to learn a good mapping from feature space to image space with identity information preserved.

\begin{table}[t]
  \caption{Identity Preserving Performance on LFW.}
  \vspace{2pt}
  \centering
  \footnotesize
  \begin{tabular}{lc}
    \toprule[1.5pt]
    Method                      & Verification Accuracy \\
    \midrule
    HPEN \cite{hpen}            & $96.25\pm0.76$        \\
    FF-GAN \cite{ffgan}         & $96.42\pm0.89$        \\
    FaceID-GAN \cite{faceidgan} & $97.01\pm0.83$        \\
    \midrule
    FaceFeat-GAN (ours)         & ${\bf 97.62\pm0.78}$  \\
    \bottomrule[1.5pt]
  \end{tabular}
  \label{tab:lfw}
\end{table}

\setlength{\tabcolsep}{3.5pt}
\begin{table}[t]
  \caption{Identity Preserving Performance on IJB-A.}
  \vspace{2pt}
  \centering
  \footnotesize
  \begin{tabular}{ccccc}
    \toprule[1.5pt]
    \multirow{2.5}{*}{\shortstack{Ratio of \\ fake data}}
               & \multicolumn{2}{c}{Verification} & \multicolumn{2}{c}{Identification} \\
    \cmidrule(l){2-5}
               &    @FIR=0.01 & @FIR=0.001     &      @Rank-1 &       @Rank-5 \\
    \midrule
    0\%        & $97.8\pm0.7$ & $93.4\pm0.5$   & $97.4\pm0.7$ & $99.1\pm0.3$  \\
    20\%       & $94.3\pm1.1$ & $86.9\pm1.1$   & $95.2\pm0.8$ & $98.5\pm0.5$  \\
    50\%       & $90.1\pm1.5$ & $79.1\pm2.7$   & $92.4\pm1.3$ & $95.7\pm0.8$  \\
    80\%       & $93.5\pm1.0$ & $85.6\pm1.6$   & $94.8\pm0.6$ & $97.7\pm0.4$  \\
    100\%      & $95.3\pm0.9$ & $90.4\pm0.8$   & $96.5\pm0.9$ & $98.4\pm0.3$  \\
    \bottomrule[1.5pt]
  \end{tabular}
  \label{tab:ijba}
\end{table}

\subsection{Image Quality}\label{subsec:image quality}
\begin{figure}[t]
  \centering
  \includegraphics[width=0.9\linewidth]{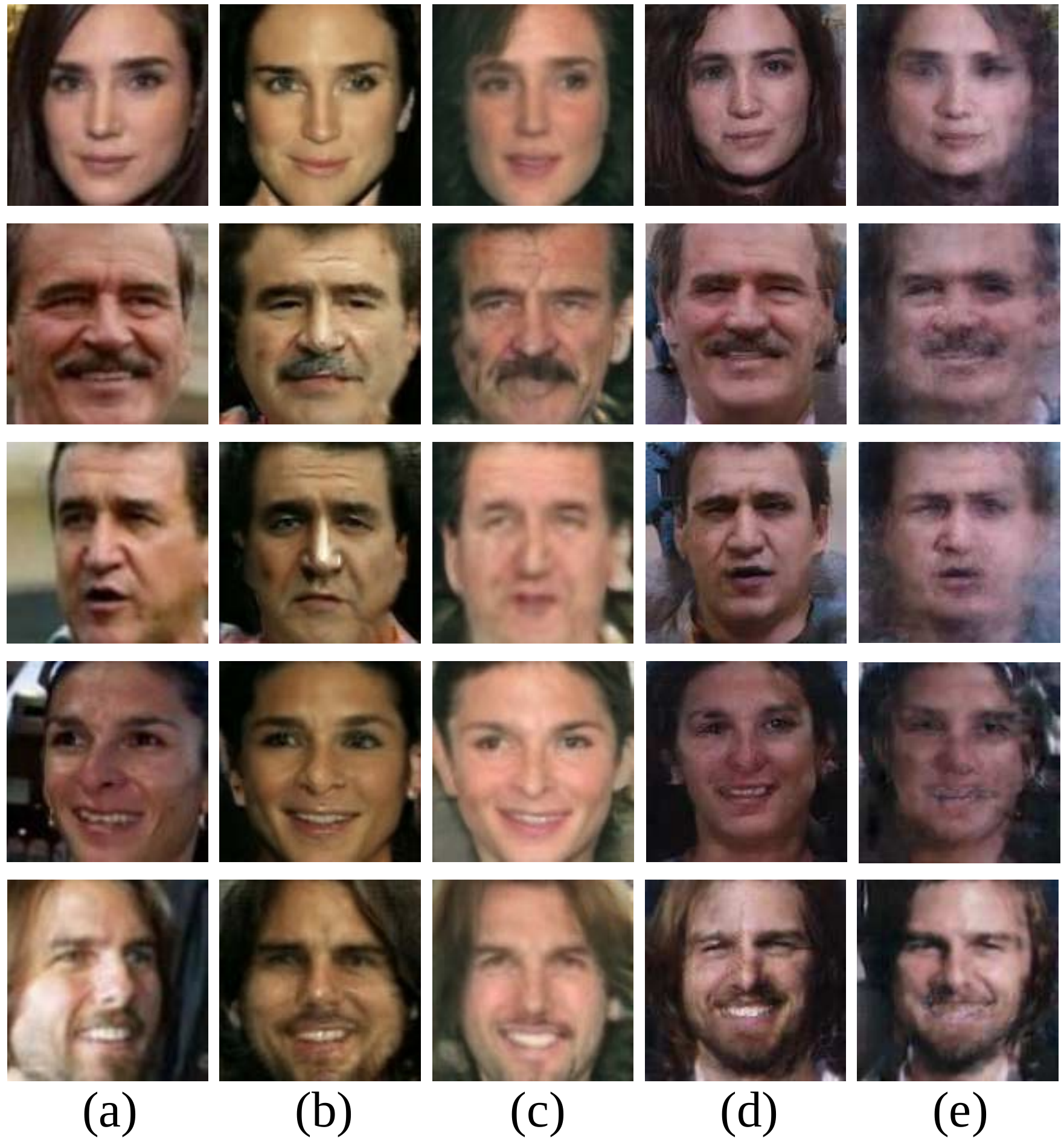}
  \caption{\small{
    Face frontalization results on LFW dataset: (a) Input, (b) FaceFeat-GAN (ours), (c) FaceID-GAN \cite{faceidgan}, (d) PIM \cite{pim}, and (e) TP-GAN \cite{tpgan}.
  }}
  \label{fig:comparison}
\end{figure}

Image quality is an important criterion to evaluate a generative model.
Fig.\ref{fig:comparison} shows some face frontalization results on LFW dataset.
Note that, FaceFeat-GAN is not designed for this task, but is able to achieve frontalization by generating the canonical pose feature.
Meanwhile, all images in Fig.\ref{fig:comparison}(b) are synthesized with the same randomly generated general feature but not the real feature extracted from the original input, which is the reason why all images are with the same illumination but different from the inputs.
This also demonstrates that FaceFeat-GAN can manipulate facial attributes independently.
All the test images are chosen by following existing work \cite{pim} but not chosen by us, leading to a fair comparison.
From Fig.\ref{fig:comparison}, we see that our methods can synthesize faces with much higher quality than prior work.

\noindent\textbf{User Study.}
A more general comparison between different identity-preserving face synthesis GANs is shown in Tab.\ref{tab:comparison}.
We randomly choose 1,000 pictures from the CelebA dataset.
By using each face as input, we produce a collection of synthesized output images with different methods.
Then we make user study on these results by asking human annotators to vote for the images with highest quality.
It turns out that our approach obtains the most votes, meaning that FaceFeat-GAN exceeds other methods in image quality.
In addition, we compute the identity similarity between each real-synthesized image pair using the independent face recognition model mentioned above, and average the results along all inputs to get an overall score for each model.
This is reported on the second column in Tab.\ref{tab:comparison}.
These two kinds of evaluations are consistent with each other, making the results reliable.
And it also demonstrates the identity-preserving ability of our method.

\subsection{Image Diversity}\label{subsec:image-diversity}
\begin{figure*}[t]
  \centering
  \includegraphics[width=0.95\linewidth]{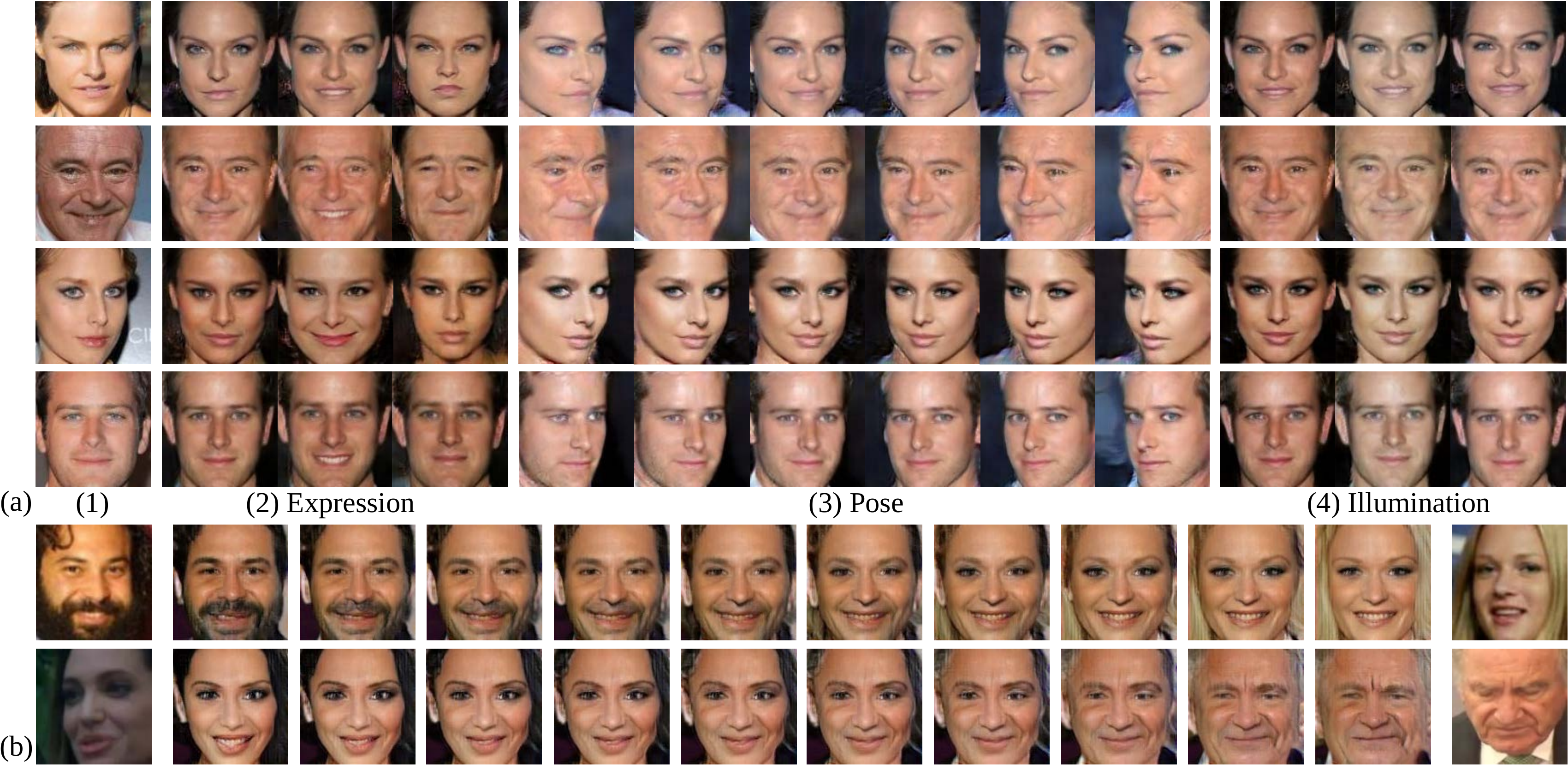}
  \caption{\small{
    (a) Highly-diverse identity-preserving face synthesis results.
    (2) and (3) are achieved by generating different 3DMM features $\f_1^s$, while images in (4) are obtained by producing various general features $\f_2^s$.
    (b) Synthesis results by interpolating identity feature $\f_{id}$.
  }}
  \label{fig:synthesis}
\end{figure*}

Besides identity-preserving property and image quality, image diversity is another merit of this work.
Some previous models including FF-GAN, TP-GAN, and PIM can only produce images with single style, such as the canonical viewpoint.
Some other methods can produce many different faces given a single input face image, but the desired face variation should be manually specified, such as the pose code in DR-GAN and 3DMM parameters in FaceID-GAN.
These methods also lack an effective supervision to decode identity information to face images, making them suffer from low image quality.
However, FaceFeat-GAN can generate face images with both high quality and high diversity, as shown in Fig.\ref{fig:synthesis}(a).
Benefiting from the novel two-stage generation process, FaceFeat-GAN is able to manipulate facial attributes, such as expression, pose and illumination, independently.
Besides generating identity-preserving faces, FaceFeat-GAN can also synthesize new faces through identity feature interpolation, as shown in Fig.\ref{fig:synthesis}(b). Interestingly we can see that the identity features also encode some other facial attributes such as the beard and age.
All results are with high quality, indicating that $G^I$ learns a good mapping from feature space, including identity feature and non-identity feature, to image space.

\setlength{\tabcolsep}{5pt}
\begin{table}[t]
  \caption{Comparison of identity similarity and image quality for different identity-preserving GANs.}
  \vspace{2pt}
  \centering
  \footnotesize
  \begin{tabular}{lccc}
    \toprule[1.5pt]
    Method & \shortstack{Similarity Score} & \shortstack{User Study Score (\%)} \\
    \midrule
    DR-GAN \cite{drgan}            & 0.548 &  4.1 \\
    FF-GAN \cite{ffgan}            & 0.592 &  7.3 \\
    TP-GAN \cite{tpgan}            & 0.625 & 11.2 \\
    Open-set GAN \cite{opensetgan} & 0.648 & 17.8 \\
    PIM \cite{pim}                 & 0.667 & 19.2 \\
    FaceID-GAN \cite{faceidgan}    & 0.653 & 18.0 \\
    \midrule
    FaceFeat-GAN (ours)            & {\bf 0.693} & ${\bf 22.4}$ \\
    \bottomrule[1.5pt]
  \end{tabular}
  \label{tab:comparison}
\end{table}

We also propose a scheme to quantitatively evaluate the diversity of the synthesized results.
With the proposed two-stage generation, we find that highly-diverse features would eventually lead to highly-diverse images.
Therefore, we define a diversity score as the average variance of each entry of the generated feature vectors.
We expect the features extracted from real images to have zero mean and unit variance because they are normalized in training.
And the diversity score of our trained model is 0.63, which is close to the real feature distribution.

\subsection{Ablation Study}\label{subsec:ablation study}
FaceFeat-GAN owns two significant improvements compared to prior work, which are
(1) two-stage generation with feature synthesis and feature-to-image mapping, and
(2) two-level competition in both image domain and feature domain.
Moreover, besides synthesizing a new image, the image generator $G^I$ also aims at reconstructing the input image.
%
In other words, there are four energy functions (components) in this work, which are $\phi_{id}$, $\phi^f$, $\phi^I$, and $\phi_{rec}$.
To validate each component of FaceFeat-GAN, we train four other models by removing one component at each time and keeping all hyper-parameters the same.
We compare these models using the metrics mentioned above, including similarity score (identity), user study score (quality), and diversity score (diversity).

Tab.\ref{tab:ablation} shows the results.
We see that (a) has much lower similarity score than (e), indicating that only learning to reconstruct the input from image domain is not enough to retain identity information.
The recognition model is essential to supervise the generating process.
The diversity score of (b) is almost 0, because the feature generator $G^f$ tends to collapse to a particular point without the feature-level competition, severely limiting the generalization ability.
Compared to (e), both (c) and (d) suffer from low image quality, demonstrating the importance of both image-level competition and pixel-wise supervision.
(c) achieves even a higher diversity score than (e), which is because feature discriminator $D^f$ is so easier to fool.
Only competing from feature level is not sufficient for $G^f$ to generate realistic features.
Therefore, we make $G^f$ to compete with $D^f$ and $D^I$ simultaneously in the full model.

\subsection{Discussion}\label{subsec:discussion}
Although FaceFeat-GAN is able to generate identity-preserving faces by balancing the trade-off between image quality and image diversity, the variation in facial expression is not high enough.
In other words, most faces are with natural or smile expressions.
There are mainly three reasons that cause the above phenomenon.
First, most faces in the training set are with such expressions.
Second, 3DMM is a parametric model. Only using it to represent expression is not good enough.
Third, the feature generators $\{G_i^f\}_{i=1}^k$ employ MLP structure, which is trivial.
Therefore, this problem should be solved with a dataset of higher diversity, a more accurate expression representation, and more carefully designed network structures.

\setlength{\tabcolsep}{5pt}
\begin{table}[t]
  \caption{Ablation study on FaceFeat-GAN.}
  \vspace{2pt}
  \centering
  \footnotesize
  \begin{tabular}{lccc}
    \toprule[1.5pt]
    Experiment Setting & \shortstack{Similarity \\ Score} & \shortstack{User Study \\ Score (\%)} & \shortstack{Diversity \\ Score}\\
    \midrule
    (a) \wo $\phi_{id}$  & 0.246 & 25.3 & 0.62 \\
    (b) \wo $\phi^f$     & 0.680 & 28.5 & 0.05 \\
    (c) \wo $\phi^I$     & 0.629 &  3.4 & 0.71 \\
    (d) \wo $\phi_{rec}$ & 0.615 &  9.6 & 0.60 \\
    \midrule
    (e) FaceFeat-GAN (full model) & {\bf 0.693} & ${\bf 33.2}$ & ${\bf 0.63}$ \\
    \bottomrule[1.5pt]
  \end{tabular}
  \label{tab:ablation}
\end{table}

\section{Conclusion}\label{sec:conclusion}
This paper presents FaceFeat-GAN, which is a novel deep generative model to achieve identity-preserving face synthesis with a two-stage synthesis procedure.
With the help of generating facial features instead of directly synthesizing faces, FaceFeat-GAN ia able to balance the trade-off between image quality and image diversity in conditional generative problem.
Extensive experimental results show the effectiveness of our proposed model.
%
%
%

{\small
\bibliographystyle{ieee}
\bibliography{ref}
}

\clearpage

\twocolumn[{
\renewcommand\twocolumn[1][]{#1}
\begin{center}
  \centering
  \includegraphics[width=0.9\linewidth]{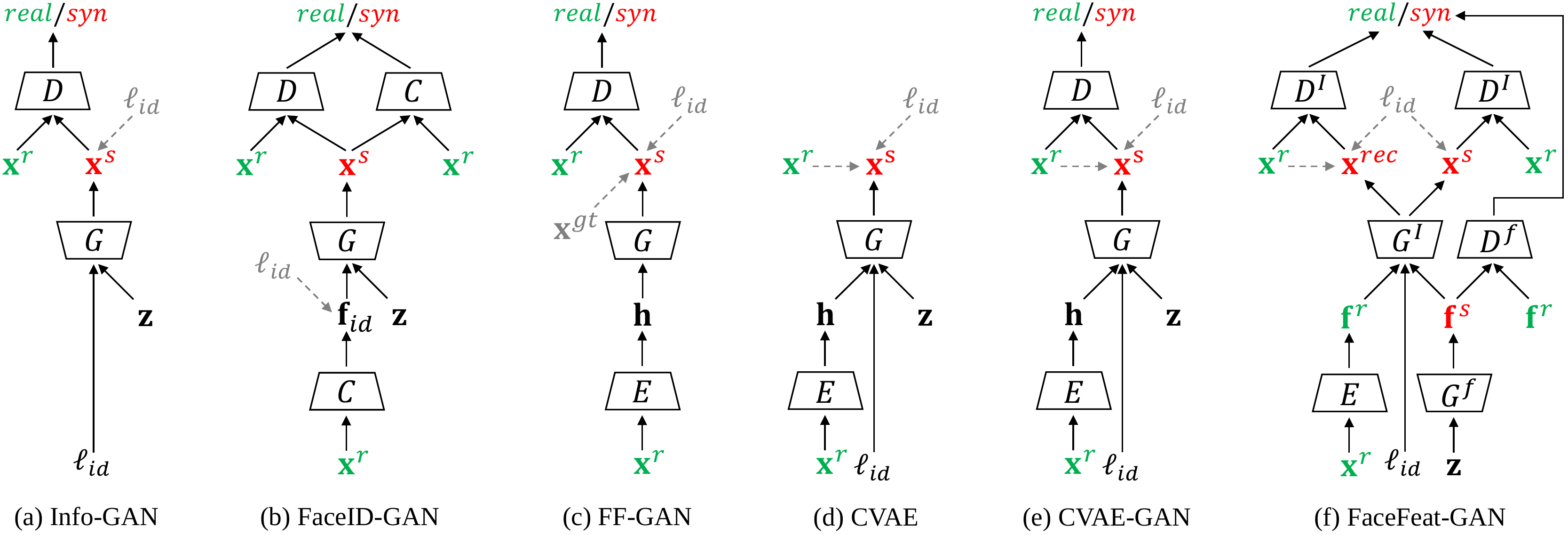}
  \captionof{figure}{\small{
    Various generative models that are designed for identity-preserving face synthesis, including (a) Info-GAN \cite{infogan}, (b) FaceID-GAN \cite{faceidgan}, (c) FF-GAN \cite{ffgan}, (d) CVAE \cite{attribute2image}, (e) CVAE-GAN \cite{cvaegan} and (f) our proposed FaceFeat-GAN.
    The \textbf{comparisons} mainly focus on (1) how to use randomness ($\z$) to improve image diversity, and (2) on how to use pixel-wise supervision ($\x^{gt}$ or $\x^r$) to improve image quality as well as identity preservation.
    Note that $\h$ is latent vector, and $\f$, discarding the superscript, is facial feature with certain semantic meaning.
    In all these figures, the {\color{black}\textbf{black arrows}} represent forward computations, whilst the {\color{gray}\textbf{grey dashed arrows}} represent backward supervisions.
    Fonts in {\color{green}\textbf{green}} and {\color{red}\textbf{red}} distinguish real and synthesized images or features respectively.
    Better viewed in color.
  }}
  \label{fig:relation}
\end{center}
}]

\begin{figure*}[t]
  \centering
  \includegraphics[width=1.0\linewidth]{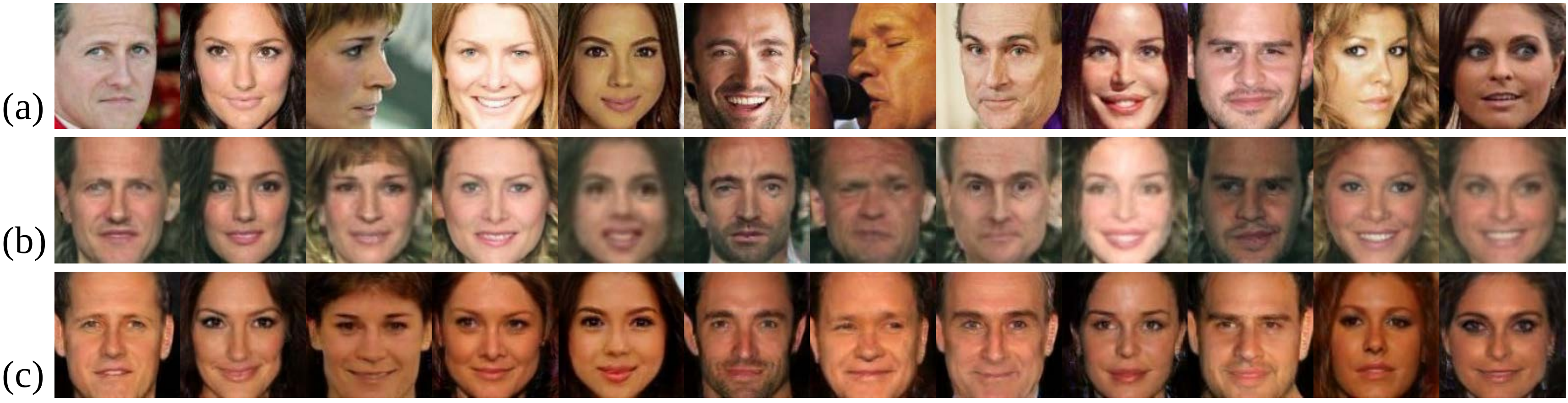}
  \caption{\small{
    Comparisons of the results synthesized by (b) FaceID-GAN \cite{faceidgan} and (c) FaceFeat-GAN, while (a) are the input images for reference.
  }}
  \label{fig:supplementary}
\end{figure*}

\section{Comparisons with Prior Work}\label{sec:comparisons}
Fig.\ref{fig:relation} illustrates the comparisons between our proposed FaceFeat-GAN and existing methods.
To preserve identity, Info-GAN \cite{infogan} passed the identity label $\ell_{id}$ to the generator $G$ and forced $G$ to output a face image $\x^s$ belonging to the desired identity by using the same label as supervision, which is shown in Fig.\ref{fig:relation}(a).
Similarly, DR-GAN \cite{drgan} replaced identity label $\ell_{id}$ with identity feature as input to provide $G$ with more information.
To further improve the identity preservation, FaceID-GAN \cite{faceidgan} in Fig.\ref{fig:relation}(b) proposed to let $G$ compete not only with the discriminator $D$, but also with the identity classifier $C$, formulating a three-player game.
Trained with the additional purpose to fool $C$, $G$ is able to better retain identity.

However, human identity is very complex and accordingly difficult to learn.
Only using identity label or identity feature as supervision is not enough to maintain identity information as much as possible.
Therefore, some work, such as FF-GAN \cite{ffgan}, introduced pixel-wise supervision, \ie use a ground truth image $\x^{gt}$ to guide the generation process, as shown in Fig.\ref{fig:relation}(c).
In this way, $G$ achieves higher-quality synthesis results by learning what pixel values to produce specifically.
Similar to this framework, TPGAN \cite{tpgan} and PIM \cite{pim} provided $G$ with both global (the entire image) and local (some key patches, \eg eyes and mouth) information.
3D-PIM \cite{3dpim} improved PIM by introducing 3D information.
Nevertheless, all of the above models are designed for face frontalization.
In other words, given an image $\x^r$, $G$ can only produce a certain output image $\x^s$ without any variation.
We can also tell that from Fig.\ref{fig:relation}(c) because $G$ does not require any randomness as input.
Furthermore, these methods rely on paired data, \ie $(\x^r, \x^{gt})$, for training, which is not easy to acquire.

In Fig.\ref{fig:relation}(d), an alternative way to solve the above problems is directly using the input image $\x^r$ as ground truth, which is presented in conditional variational auto-endocder (CVAE) \cite{attribute2image}.
But the synthesized image $\x^s$ shows blurring due to the lack of the competition between $G$ and $D$.
CVAE-GAN \cite{cvaegan} involved GAN into CVAE to get advantages from both models.
However, in Fig.\ref{fig:relation}(e), $\x^r$ is used to supervise $\x^s$ no matter what the input noise $\z$ is.
This will cause confusion to $G$ and severely limit the diversity of synthesized results.
\cite{opensetgan} tried to use different attribute images as supervisions instead of $\x^r$, but there is no guarantee that attribute image has same identity as input, which will lead to identity information loss.

In contrast, we propose FaceFeat-GAN in Fig.\ref{fig:relation}(f) to balance the trade-off between \textbf{image quality} and \textbf{image diversity}.
This goal is achieved with a two-stage generation.
This first stage $G^f$ accounts for diversity by producing facial features $\f^s$ with large variety from random vector $\z$.
To ensure the realness of generated features, we employ feature discriminator $D^f$ to differentiate real and synthesized domains from feature space.
The second stage $G^I$ renders photo-realistic identity-preserving face images by taking $\f^s$ and $\ell_{id}$ as inputs.
We also have $D^I$ to compete with $G^I$ from image space.
To learn a better mapping from feature space to image space, we introduce pixel-wise supervision as shown in Fig.\ref{fig:relation}(e).
Specially, $G^I$ not only generates a new image $\x^s$, but also produces $\x^{rec}$ to reconstruct the input image $\x^r$.
This novel two-stage design resolves the contradiction between how to introduce randomness to improve image diversity and how to apply per-pixel supervision to improve image quality.

\section{More Results}\label{sec:results}
FaceFeat-GAN can generate identity-preserving face images with both high quality and high diversity.
Fig.\ref{fig:supplementary} shows some comparisons between our proposed FaceFeat-GAN and prior work that is capable of synthesized highly-diverse images, \ie FaceID-GAN \cite{faceidgan}.
We can tell that the results produced by FaceFeat-GAN are much more photo-realistic and better preserve identity.
This benefits from the pixel-wise supervision introduced in the second stage.
Meanwhile, for each input image, we generate a series of images by controlling $\z_1$ and $\z_2$, to demonstrate that FaceFeat-GAN can not only improve image quality, but also generate images with large variety, outperforming previous face frontalization methods \cite{ffgan, pim, 3dpim, tpgan}.
This is due to the help of the feature generators in the first stage.
Please check the demo on YouTube. \url{https://youtu.be/4yqYGCCXWbM}

In summary, the novel two-stage generation fairly balances the trade-off between image quality and image diversity in conditional generative problem.

\end{document}